# Disambiguating bilingual nominal entries against WordNet

German Rigau.*
Departament de Llenguatges i Sistemes Informàtics. Universitat Politècnica de Catalunya.
Pau Gargallo 5, 08028 Barcelona. Spain. g.rigau@lsi.upc.es

Eneko Agirre.**
Lengoaia eta Sistema Informatikoak Saila. Euskal Herriko Unibertsitatea.
p.k. 649, 20080 Donostia. Spain. jibagbee@si.ehu.es

## 1. INTRODUCTION

One reason why the lexical capabilities of NLP systems have remained weak is because of the labour intensive nature of encoding lexical entries for the lexicon. It has been estimated that the average time needed to construct manually a lexical entry for a Machine Translation system is about 30 minutes [Neff et al. 93]. The automatic acquisition of lexical knowledge is the main field of the research work presented here. In particular, this paper explores the acquisition of conceptual knowledge from bilingual dictionaries (French/English, Spanish/English and English/Spanish) using a pre-existing broad coverage Lexical Knowledge Base (LKB) WordNet [Miller 90].

The automatic acquisition of lexical knowledge from monolingual machine-readable dictionaries (MRDs) has been broadly explored (e.g. [Boguraev & Briscoe 90], [Artola 93], [Castellón 93], [Wilks et al. 93], [Dolan et al. 93]), while less attention has been paid to bilingual dictionaries (e.g. [Ageno et al. 94], [Knight & Luk 94]).

Bilingual dictionaries contain information about the connection of vocabularies in two different languages. However, MRDs are made for human readers and the information contained in it is not immediately usable as a computational lexicon. For instance word translations are not marked with a sense or group of senses (sense mismatch problem), but they are sometimes annotated with subject field codes or cue words in the source language.

Two different, complementary approaches are explored in this paper. Both of them use WordNet to obtain a multilingual LKB (MLKB). The resulting MLKB has the same structure as WordNet, but some nodes are attached additionally to disambiguated vocabulary of other languages.

In one of the approaches each entry of the dictionary is taken in turn, exploiting the information in the entry itself. The inferential capability for disambiguating the translation is given by Semantic Density over WordNet [Agirre & Rigau, 95]. In the other approach, the bilingual dictionary was merged with WordNet, exploiting mainly synonymy relations. Each of the approaches was used in a different dictionary. The first approach was used on a French-English dictionary (using one direction only), and the second approach on a Spanish-English/English-Spanish dictionary (both directions).

* German Rigau was supported by a grant from the Ministerio de Educación y Ciencia.
** Eneko Agirre was supported by a grant from the Basque Government.

After this short introduction, section 2 shows some experiments and results using Semantic Density on the bilingual French/English dictionary. In section 3 several complementary techniques and results using the Spanish bilingual dictionaries are explained.

## 2. WORD SENSE DISAMBIGUATION USING CONCEPTUAL DENSITY

2.1 The French/English bilingual dictionary

The French/English bilingual dictionary contains 21,322 entries. Each entry can comprise several or a single sense of the source word, which in the scope of this paper we will call subentries. For instance, the entry for 'maintien' is split in two subentries:

```
maintien n.m. (attitude) bearing; (conservation) maintenance.

maintien 1: n.m. (attitude) bearing
maintien 2: n.m. (conservation) maintenance
```

The dictionary has 31,502 such subentries, from which 16,917 are nominal subentries.

Each subentry can have the following fields: part of speech (always), semantic field (one out of a set of 20, e.g. `comm.` in `trésor 2` in the example below), cue in French (e.g. `ressources` in `trésor 2`) and one or several translations in English (always). The semantic field and the cue in French are used to determine the context or the usage of the French word when translated by the subentry.

```
folie 1: n.f. madness
provision 1: n.f. supply, store
trésor 2: n.m. (ressources) (comm.) finances
```

In order to figure out which WordNet sense(s) fit(s) best the French headword, the algorithm needs contextual information (as we humans do). If we do not have any contextual information, and the translation has more than one sense, it is not possible to find the correct sense(s)[1]. The cases where we can try to disambiguate the translation are the following:

1) one of the translation words is monosemous in WordNet
2) the translation is given by a list of words
3) a cue in French is provided alongside the translation
4) a semantic field is provided

From the examples above, `folie`'s translation has more than one sense and therefore is not a member of any of the cases. `provision` has two translation polysemous translations and therefore belongs to case 2. `trésor` has a monosemous translation and also comes with a French cue (`ressources`) and a semantic field (`comm` meaning commercial), and therefore belongs to cases 2, 3 and 4.

The figures for combinations of the above cases found in the bilingual dictionaries are the following:

---

[1] In this work we try to assign a single sense to the translations.

| | | |
|---|---:|---:|
| translation not in WordNet | 4,081 | 24% |
| unique translation, n senses | 4,761 | 28% |
| any combination of cases 1,2,3,4 | 8,075 | 48% |
| total | 16,917 | 100% |

Table 1

The figures mean that, from all the senses of French nouns, we can disambiguate at most 48% of them. The coverage of WordNet is not very impressive, only 76% of the English nouns in the bilingual dictionary. This is caused by several problems that will be dealt with below.

The bilingual subentries that provide disambiguation information have the distribution shown below. Some subentries belong at the same time to more than one case.

| | | |
|---|---:|---:|
| case 1; 1 sense | 5,039 | 30% |
| case 2; more than one translation | 630 | 4% |
| case 3; cue in French | 2,954 | 17% |
| case 4; semantic field | 1,067 | 6% |

Table 2

Those that have a monosemous unique translation can be directly linked. Besides we still have not experimented with the use of semantic fields. Therefore, the algorithm will focus on bilingual subentries with multiple translations and/or cues in French.

2.2 Treatment of complex translations and cues

In the previous paragraph, it was said that 24% of the translations were not found in WordNet. A quick look at some of the translations revealed that the failure was sometimes caused by the translation being in a plural form, being composed by a whole noun phrase, brackets, etc. The same situation was observed in the cues, which were often composed by a phrase or a list of phrases. We call these translations and cues *complex*. Some examples of complex translations and cues follow:

```
batterie 2: n.f. (mus.) drums
e'poux 2: n.m. the married couple
escale 2: n.f. (port) port of call
microplaquette 1: n.f. (micro) chip
remonte'e 2: n.f. (d'eau, de prix) rise
```

The treatment for the translations and cues that could not be found directly in WordNet or the bilingual dictionary respectively was done in two steps. First, a morphological analysis was performed, and if it was not successful, combinations of the component words were tried.

A) morphological analysis: For English we use the morphological analyser provided by WordNet. In the case of French, a naive morphological analysis is tried (valid for nouns only), checking the resulting potential lemmas against the bilingual dictionary itself. For instance, morphological lookup for the translation for `batterie 2` would yield `drum`.

B) complex phrases: when the translation or cue is composed by more than one word, several combinations of the component words are tried. The longest combination of words that is successfully looked-up is returned. If no combination is succesful, then all the component words that are correct nouns (according to WordNet for English, and the bilingual dictionary for French) are returned. For the translation of `e'poux 2` this procedure would return `married couple`, which is correctly found in WordNet. In another example, `port of call` would yield both `port` and `call`. The same applies for cues: the processing of the cue `d'eau, de prix` would output both `eau` and `prix`. Brackets are also taken into account, but in this case the words inside brackets would never be returned on their own, only as components of a compound noun.

A sample of 50 complex translations was evaluated, to see the reliability of the method proposed. In 21% of the results, the single correct translation was proposed. The most significant part of the translation was captured in 67% of the cases, and only 12% of the proposed translations were wrong.

After processing the English translations, it was found that the coverage of WordNet increased from 76% to 95%, leaving only 891 subentries that could not be processed. This means that the figures for all cases in tables 1 and 2 change, as shown in tables 1' and 2'.

| translation not in WordNet | 891 | 5% |
|---|---:|---:|
| unique translation, n senses | 6,440 | 38% |
| any combination of cases 1,2,3,4 | 9,586 | 57% |
| total | 16,917 | 100% |

Table 1'

| case 1; 1 sense | 5,119 | 30% |
|---|---:|---:|
| case 2; more than one translation | 958 | 6% |
| case 3; cue in French | 3,702 | 22% |
| case 4; semantic field | 1,365 | 8% |

Table 2'

## 2.3 The disambiguation procedure

In the core of the disambiguation procedure we use conceptual density as described in [Agirre & Rigau, 95], [Rigau 94] and [Agirre et al. 94]. Conceptual Density provides a basis for determining relatedness among words, taking as reference a structured hierarchical net which in this case is WordNet. For instance, in figure 1 we have a word W with four senses. Each sense belongs to a subtree in the hierarchical net. The dots in the subtrees represent the senses of either the word to be disambiguated (W) or the words in the context. Semantic Density will yield the highest density for the subtree containing more senses of those, relative to the total amount of senses in the subtree.

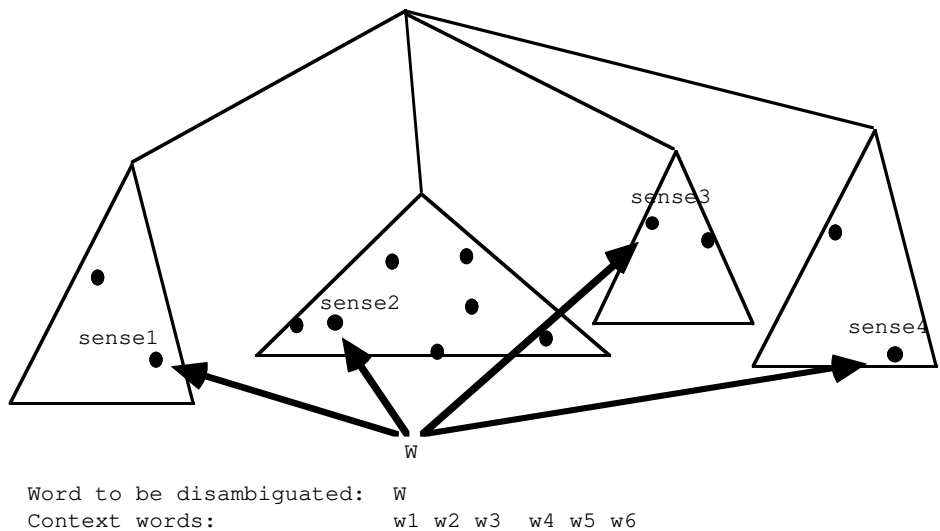

```
Word to be disambiguated:  W
Context words:             w1 w2 w3  w4 w5 w6
```
Figure 1: senses of a word in WordNet

The relatedness of a certain word-sense to the words in the context allows us to select that sense over the others. Following with the example in figure 1, sense2 would be chosen for W, because it belongs to the subtree with highest Semantic Density. In some cases more than one sense of the word to be disambiguated will belong to the selected subtree. In that case multiple senses are returned.

The context words are provided by the cue words in French and multiple translations. Cue words are in French, and therefore need to be translated into English, which is done using the bilingual dictionary.

In order to evaluate the contribution of each kind of contextual information separately, two experiments where performed on two sets of subentries: a set comprising French cues with a single translation word, and a set containing more than one translation but without any French cue.

2.4 Estimate the contribution of French cues

French cues are looked up in the bilingual dictionary, and all the English translations of the cue are input to the algorithm alongside the English translation. These English words will provide the necessary contextual information for the disambiguation of the translation.

A set of experiments was performed to evaluate the expected precision when disambiguating subentries that had a single English translation and a French cue. For this purpose, 59 French subentries fulfilling the given condition were selected at random

The precision and coverage are shown in the second line of the table below. The precision is considerably higher than random guessing[2]. The error rate was deemed too high, specially for some of the potential applications. In order to reduce the error rate several heuristics were tried. Declining to disambiguate translations with more than 5 senses was the most successful. As the third line of the following table shows, precision raised at the cost of the coverage.

---

[2] The figure for random guessig takes into account all noun entries. It was obtained analytically using the polysemy figures for all translations.

|  | precision | coverage |
|---|---|---|
| random guessing | 44.8% | - |
| original results | 67.4% | 72.9% |
| heuristic | 83.3% | 50.8% |

Table 3

2.5 Estimate contribution of several translations

In this experiment 30 subentries that had more than one English translation were selected at random. The disambiguation algorithm was fed with the set of translation words and produced a set of WordNet synsets. The results, with and without applying the heuristic, are the following:

|  | precision | coverage |
|---|---|---|
| random guessing | 44.8% | - |
| original results | 89.3% | 93.3% |
| heuristic | 90.9% | 73.3% |

Table 4

Performance for this subset of the definitions is considerably better than for French cues. The heuristic does not yield significant improvement in precision, and the original results are preferred.

2.6 Overall results

Table 5 summarises the overall results. The algorithm was run over all the subentries, except those containing semantic fields. This means that in the best case, 8,221[3] subentries (53% of the total 15,552) could be linked. For a given subentry, whether it was monosemous or not was checked first. If not, disambiguation using multiple translations was tried, and last, cues in French were used. Monosemous translations account for most of the links made. The low coverage when disambiguating with French cues accounts for most of the failures to make links.

| no result | 8,311 | 53% |
|---|---|---|
| result obtained | 7,241 | 47% |
| case 1; 1 sense | 5,119 | 33% |
| case 2; >1 trans | 723 | 5% |
| case 3; cue | 1,399 | 9% |
| total | 15,552 | 100% |

Table 5

The links made, as calculated in the previous experiments, are highly reliable. The confidence for monosemous links (case 1) would be 100% if it not were because of complex translations, for which 88% of precision can be expected. For case 2, 93% of correct answers can be expected which descends to 83% for case 3 subentries.

Overall coverage of this method will hopefully improve when semantic fields are taken into account.

---

[3] Calculated from tables 1' and 2', substracting the number of semantic fields from the overall combination of cases 1,2,3 and 4.

## 3. MERGING LEXICAL KNOWLEDGE RESOURCES

Four experiments have been performed exploiting simple properties to attach Spanish nouns from the Spanish/English-English/Spanish bilingual dictionary to noun synsets in WordNet 1.5.

The nominal part of WordNet 1.5 has 60557 synsets and 87642 English nouns (76127 monosemous). The Spanish/English bilingual dictionary contains 12370 Spanish nouns and 11467 English nouns in 19443 connections among them. On the other hand, the English/Spanish bilingual dictionary is less informative than the other one containing only 10739 English nouns, 10549 Spanish nouns in 16324 connections.

Merging both dictionaries a list of equivalence pairs of nouns have been obtained. The combined dictionary contains 15848 English nouns, 14880 Spanish nouns and 28131 connections.

For instance, for the word "masa" in Spanish the following list of equivalence pairs can be obtained:

```
------------------------   English/Spanish
bulk   masa
dough  masa
mass   masa
------------------------   Spanish/English
cake   masa
crowd_of_people   masa
dough   masa
ground  masa
mass   masa
mortar  masa
volume  masa
```

From the combined dictionary, there are only 12665 English nouns placed in WordNet 1.5 which represents 19383 synsets. That is, the maximum coverage we can expect of WordNet1.5 using both bilingual Spanish/English dictionaries is 32%. In the next table the summarised amount of data is shown.

|  | English nouns | Spanish nouns | synsets | connections |
|---|---|---|---|---|
| WordNet1.5 | 87,642 | - | 60,557 | 107,424 |
| Spanish/English | 11,467 | 12,370 | - | 19,443 |
| English/Spanish | 10,739 | 10,549 | - | 16,324 |
| Merged Bilingual | 15,848 | 14,880 | - | 28,131 |
| Maximum Coverage | 12,665 | 13,208 | 19,383 | 24,613 |
| of WordNet | 14% | - | 32% | - |
| of bilingual | 80% | 90% | - | 87% |

Table 6

The connection of Spanish nouns to Synsets in WordNet 1.5 has been performed in the following cases:

1) Those Spanish nouns translations of monosemous English nouns (one sense in WordNet). Considering for instance that the noun abduction has only one sense in WordNet1.5[4] :

> Synonyms/Hypernyms (Ordered by Frequency) of noun abduction
> 1 sense of abduction
>
> Sense 1
> <abduction>
>     => <capture, seizure>
>      => <felony>
>         => <crime, law-breaking>
>          => <evildoing, transgression>
>           => <wrongdoing, misconduct>
>           => <activity>
>                 => <act, human action, human activity>

and there are two possible translations for abduction for Spanish

> secuestro   <-->   abduction
> rapto       <-->   abduction

the following attachment has been produced:

> <abduction>  <-->   <secuestro, rapto>

Only 6616 English nouns from the equivalence pairs list are monosemous (42% of the total English nouns). Thus, this simple approach has produced 9057 connections among 7636 Spanish nouns and 5963 synsets of WordNet1.5 with a very high degree of confidence. The polysemous degree in this case is 1.19 synsets per Spanish noun with 1.52 Spanish nouns per synset. Next table shows the results following this process.

|                  | English nouns | Spanish nouns | synsets | connec. | Poly. | Syn. |
|------------------|---------------|---------------|---------|---------|-------|------|
| WordNet          | 87,642        | -             | 60,557  | 107,424 | 1.2   | 1.8  |
| Bilingual        | 15,848        | 14,880        | -       | 28,131  |       |      |
| Maximum Coverage | 12,665        | 13,208        | 19,383  | 24,613  | 1.9   | 1.3  |
| Case 1           | 6,616         | **7,636**     | 5,963   | 9,057   | 1.2   | 1.5  |
| of WordNet       | 8%            | -             | 10%     | -       |       |      |
| of Bilingual     | 42%           | 51%           | -       | -       |       |      |
| of Maximum       | 52%           | 58%           | 30%     | 37%     |       |      |
| of total         | 58%           | 63%           | 37%     | 37%     |       |      |
| Total            | 11,470        | 12,039        | 15,897  | 24,535  |       |      |

Table 7

2) Those Spanish nouns with only one translation (although, the translation could be polysemous). Consider for instance the only translation found into the merged dictionary for the Spanish noun *anfibio* :

> amphibian   <-->   anfibio

---

[4] In the following examples, brackets are used indicating synsets (concepts) and => means hyponym-of.

This process has produced three possible connections for the English WordNet1.5 amphibian:

```
<amphibian, amphibious vehicle>     <-->    <anfibio>
<amphibian, amphibious aircraft>    <-->    <anfibio>
<amphibian>                         <-->    <anfibio>
    => <vertebrate, craniate>
```

There are 8524 Spanish nouns with only one translation. These Spanish nouns are equivalence candidates of 7507 English nouns but only 6066 of these are present in WordNet1.5. Thus, this approach has generated 14164 connections among 7000 Spanish nouns and 10674 synsets. The polysemous ratio is 2.02 synsets per Spanish noun and there are 1.33 Spanish word per synset. In the following table the results for this approach are shown.

|                  | English nouns | Spanish nouns | synsets | connec. | Poly. | Syn. |
|------------------|---------------|---------------|---------|---------|-------|------|
| WordNet          | 87,642        | -             | 60,557  | 107,424 | 1.2   | 1.8  |
| Bilingual        | 15,848        | 14,880        | -       | 28,131  |       |      |
| Maximum Coverage | 12,665        | 13,208        | 19,383  | 24,613  | 1.9   | 1.3  |
| Case 2           | 6,066         | 7,000         | **10,674** | **14,164** | 2.0 | 1.3 |
| of WordNet       | 7%            | -             | 18%     | -       |       |      |
| of Bilingual     | 38%           | 47%           | -       | -       |       |      |
| of Maximum       | 48%           | 53%           | 55%     | 58%     |       |      |
| of total         | 53%           | 58%           | 67%     | 58%     |       |      |
| Total            | 11,470        | 12,039        | 15,897  | 24,535  |       |      |

Table 8

3) Those English nouns (although, the translation could be polysemous) with only one translation. Consider the unique translation of banishment for the nominal part of the bilingual dictionaries:

```
banishment <--> destierro
```

Thus, the Spanish noun *destierro* has been attached to both synsets of banishment in WordNet:

```
<banishment, ostracism>    <-->    <destierro>
    => <exclusion>
        => <situation, state of affairs>
        => <state>

<banishment, proscription>    <-->    <destierro>
    => <rejection>
        => <act, human action, human activity>
```

There are 10285 English nouns with only one translation (out of 7383 are present in WordNet). These English nouns are equivalence translations of 8556 Spanish nouns. In this case, 11089 connections have been produced among 6470 Spanish nouns and 10223 synsets. Thus, the polysemous ratio is 1.71 synsets per Spanish noun with 1.08 Spanish noun per synset. In next table this data is summarized.

|                  | English nouns | Spanish nouns | synsets | connec. | Poly. | Syn. |
|------------------|---------------|---------------|---------|---------|-------|------|
| WordNet          | 87,642        | -             | 60,557  | 107,424 | 1.2   | 1.8  |
| Bilingual        | 15,848        | 14,880        | -       | 28,131  |       |      |
| Maximum Coverage | 12,665        | 13,208        | 19,383  | 24,613  | 1.9   | 1.3  |
| Case 3           | **7,383**     | 6,470         | 10,223  | 11,089  | 1.7   | 1.1  |
| of WordNet       | 8%            | -             | 17%     | -       |       |      |
| of Bilingual     | 47%           | 44%           | -       | -       |       |      |
| of Maximum       | 58%           | 49%           | 53%     | 45%     |       |      |
| of total         | 64%           | 54%           | 64%     | 45%     |       |      |
| Total            | 11,470        | 12,039        | 15,897  | 24,535  |       |      |

Table 9

4) Those synsets with several English nouns with the same translation. Consider the following translations for the word *error* in the merged bilingual dictionary:

```
error    <-->   error
mistake  <-->   error
```

then this process can generate the following attachment:

```
<mistake, error, fault>    <-->   <error>
    => <failure>
        => <nonaccomplishment, nonachievement>
            => <act, human action, human activity>

<error, mistake>           <-->   <error>
    => <misstatement>
        => <statement>
            => <message, content, subject matter, substance>
                => <communication>
                    => <social relation>
                        => <relation>
                            => <abstraction>
```

In this case, 3164 connections among 2261 Spanish nouns and 2195 synsets have been found. That means a polysemous ratio of 1.40 synsets per Spanish noun and 1.44 Spanish nouns per synset. The next table summarises the last approach.

|                  | English nouns | Spanish nouns | synsets | connec. | Poly. | Syn. |
|------------------|---------------|---------------|---------|---------|-------|------|
| WordNet          | 87,642        | -             | 60,557  | 107,424 | 1.2   | 1.8  |
| Bilingual        | 15,848        | 14,880        | -       | 28,131  |       |      |
| Maximum Coverage | 12,665        | 13,208        | 19,383  | 24,613  | 1.9   | 1.3  |
| Case 4           | 2,092         | 2,261         | 2,195   | 3,164   | 1.4   | 1.4  |
| of WordNet       | 2%            | -             | 4%      | -       |       |      |
| of Bilingual     | 13%           | 15%           | -       | -       |       |      |
| of Maximum       | 17%           | 17%           | 11%     | 13%     |       |      |
| of total         | 18%           | 19%           | 14%     | 13%     |       |      |
| Total            | 11,470        | 12,039        | 15,897  | 24,535  |       |      |

Table 10

Merging all the connections we have obtained a micro-Spanish WordNet (with errors). The resulting data has 24535 connections among 12039 Spanish nouns and 15897 synsets of WordNet1.5. That is to say, a polysemous ratio of 2.03 synsets per Spanish noun with 1.54 synonymy degree. The next table shows the overall data:

| | English nouns | Spanish nouns | synsets | connec. | Poly. | Syn. |
|---|---|---|---|---|---|---|
| WordNet | 87,642 | | 60,557 | 107,424 | 1.2 | 1.8 |
| Bilingual | 15,848 | 14,880 | | 28,131 | | |
| Maximum Coverage | 12,665 | 13,208 | 19,383 | 24,613 | 1.9 | 1.3 |
| Case 1 | 6,616 | **7,636** | 5,963 | 9,057 | 1.2 | 1.5 |
| Case 2 | 6,066 | 7,000 | **10,674** | **14,164** | 2.0 | 1.3 |
| Case 3 | **7,383** | 6,470 | 10,223 | 11,089 | 1.7 | 1.1 |
| Case 4 | 2,092 | 2,261 | 2,195 | 3,164 | 1.4 | 1.4 |
| Total | 11,470 | 12,039 | 15,897 | 24,535 | 2.0 | 1.5 |
| of WordNet | 13% | - | 26% | - | | |
| of Bilingual | 72% | 80% | - | - | | |
| of Maximum | 90% | 91% | 82% | 100% | | |

Table 11

We have tested manually one hundred connections. 78 out of 100 were correct. Obviously, the most productive cases are the cases that introduce more errors.

## 4. CONSIDERATIONS

This paper shows that disambiguating bilingual nominal entries, and therefore linking bilingual dictionaries to WordNet is a feasible task. The complementary approaches presented here, Semantic Density on entry information and merging taking profit of dictionary structure, both attain high levels of precision on their own. The combination of both techniques, alongside using the semantic fields left aside by the first approach, should yield better precision and a raise in coverage. For instance, the first approach focuses on the information in the French/English direction of the dictionary, without using the reverse direction or exploiting the structure of the dictionary as in the second approach. The second approach, on the other hand, could take profit from both the information in each entry and the inferential capability of Semantic Density.